\documentclass[sigconf]{acmart}

\usepackage{pifont}
\usepackage{cleveref}
\usepackage{caption}
\usepackage{threeparttable}
\usepackage{arydshln}
\usepackage{multirow}
\usepackage{array}
\usepackage{rotating}
\usepackage{colortbl}
\usepackage{booktabs}
\usepackage{graphicx}
\usepackage{subcaption}

\crefformat{section}{\S#2#1#3}
\crefformat{subsection}{\S#2#1#3}
\crefformat{subsubsection}{\S#2#1#3}
\crefrangeformat{section}{\S\S#3#1#4 to~#5#2#6}
\crefmultiformat{section}{\S\S#2#1#3}{ and~#2#1#3}{, #2#1#3}{ and~#2#1#3}

\AtBeginDocument{%
  \providecommand\BibTeX{{%
    \normalfont B\kern-0.5em{\scshape i\kern-0.25em b}\kern-0.8em\TeX}}}

\setcopyright{acmcopyright}

\copyrightyear{2024} 
\acmYear{2024} 
\setcopyright{acmlicensed}\acmConference[WSDM '24]{Proceedings of the 17th ACM International Conference on Web Search and Data Mining}{March 4--8, 2024}{Merida, Mexico}
\acmBooktitle{Proceedings of the 17th ACM International Conference on Web Search and Data Mining (WSDM '24), March 4--8, 2024, Merida, Mexico}
\acmPrice{15.00}
\acmDOI{10.1145/3616855.3635753}
\acmISBN{979-8-4007-0371-3/24/03}

\begin{document}

\title{Hierarchical Multimodal Pre-training for Visually Rich Webpage Understanding}



\author{Hongshen Xu}
\orcid{0000-0002-6770-6564}
\email{xuhongshen@sjtu.edu.cn}
\affiliation{
\department{X-LANCE lab}
\institution{Shanghai Jiao Tong University}
\city{Shanghai}
\country{China}}

\author{Lu Chen}
\orcid{0000-0001-8687-4806}
\email{chenlusz@sjtu.edu.cn}
\affiliation{
\department{X-LANCE lab}
\institution{Shanghai Jiao Tong University}
\city{Shanghai}
\country{China}}
\authornote{Lu Chen and Kai Yu are the corresponding authors.}

\author{Zihan Zhao}
\orcid{0009-0000-6832-8107}
\email{zhao_mengxin@sjtu.edu.cn}
\affiliation{
\department{X-LANCE lab}
\institution{Shanghai Jiao Tong University}
\city{Shanghai}
\country{China}}

\author{Da Ma}
\orcid{0009-0009-6679-2032}
\email{mada123@sjtu.edu.cn}
\affiliation{
\department{MoE Key Lab of Artificial Intelligence}
\institution{Shanghai Jiao Tong University}
\city{Shanghai}
\country{China}}

\author{Ruisheng Cao}
\orcid{0000-0003-4635-4368}
\email{211314@sjtu.edu.cn}
\affiliation{
\department{MoE Key Lab of Artificial Intelligence}
\institution{Shanghai Jiao Tong University}
\city{Shanghai}
\country{China}}

\author{Zichen Zhu}
\orcid{0000-0002-9321-5662}
\email{JamesZhutheThird@sjtu.edu.cn}
\affiliation{
\department{MoE Key Lab of Artificial Intelligence}
\institution{Shanghai Jiao Tong University}
\city{Shanghai}
\country{China}}

\author{Kai Yu}
\orcid{0000-0002-7102-9826}
\email{kai.yu@sjtu.edu.cn}
\authornotemark[1]

\affiliation{
\department{MoE Key Lab of Artificial Intelligence}
\institution{Shanghai Jiao Tong University}
\city{Shanghai}
\country{China}}

\renewcommand{\shortauthors}{Hongshen Xu et al.}

\begin{abstract}
The growing prevalence of visually rich documents, such as webpages and scanned/digital-born documents (images, PDFs, etc.), has led to increased interest in automatic document understanding and information extraction across academia and industry. Although various document modalities, including image, text, layout, and structure, facilitate human information retrieval, the interconnected nature of these modalities presents challenges for neural networks. In this paper, we introduce WebLM, a multimodal pre-training network designed to address the limitations of solely modeling text and structure modalities of HTML in webpages. Instead of processing document images as unified natural images, WebLM integrates the hierarchical structure of document images to enhance the understanding of markup-language-based documents. Additionally, we propose several pre-training tasks to model the interaction among text, structure, and image modalities effectively. Empirical results demonstrate that the pre-trained WebLM significantly surpasses previous state-of-the-art pre-trained models across several webpage understanding tasks. The pre-trained models and code are available at https://github.com/X-LANCE/weblm.

\end{abstract}




\maketitle

\section{Introduction}

Visually rich documents have become the primary means of organizing, presenting, storing, and transmitting information over the Internet for billions of individuals. Recent advancements in the domain of deep learning and natural language processing have led to increasing attention toward automatically understanding and extracting information from these documents due to their diverse application scenarios \cite{jaume2019funsd, huang2019icdar2019, park2019cord, gralinski2020kleister, mathew2021docvqa}. To address the challenges posed by the cross-modality interconnections within visually rich documents, self-supervised training on large-scale unlabeled data \cite{devlin2018bert,liu2019roberta,clark2020electra} and multimodal pre-training techniques \cite{radford2021learning,tan2019lxmert,lu2019vilbert,chen2020uniter} have emerged as promising approaches for Visually Rich Document Understanding (VRDU) tasks.


\begin{figure}[t]
\centering\includegraphics[width=0.85\linewidth]{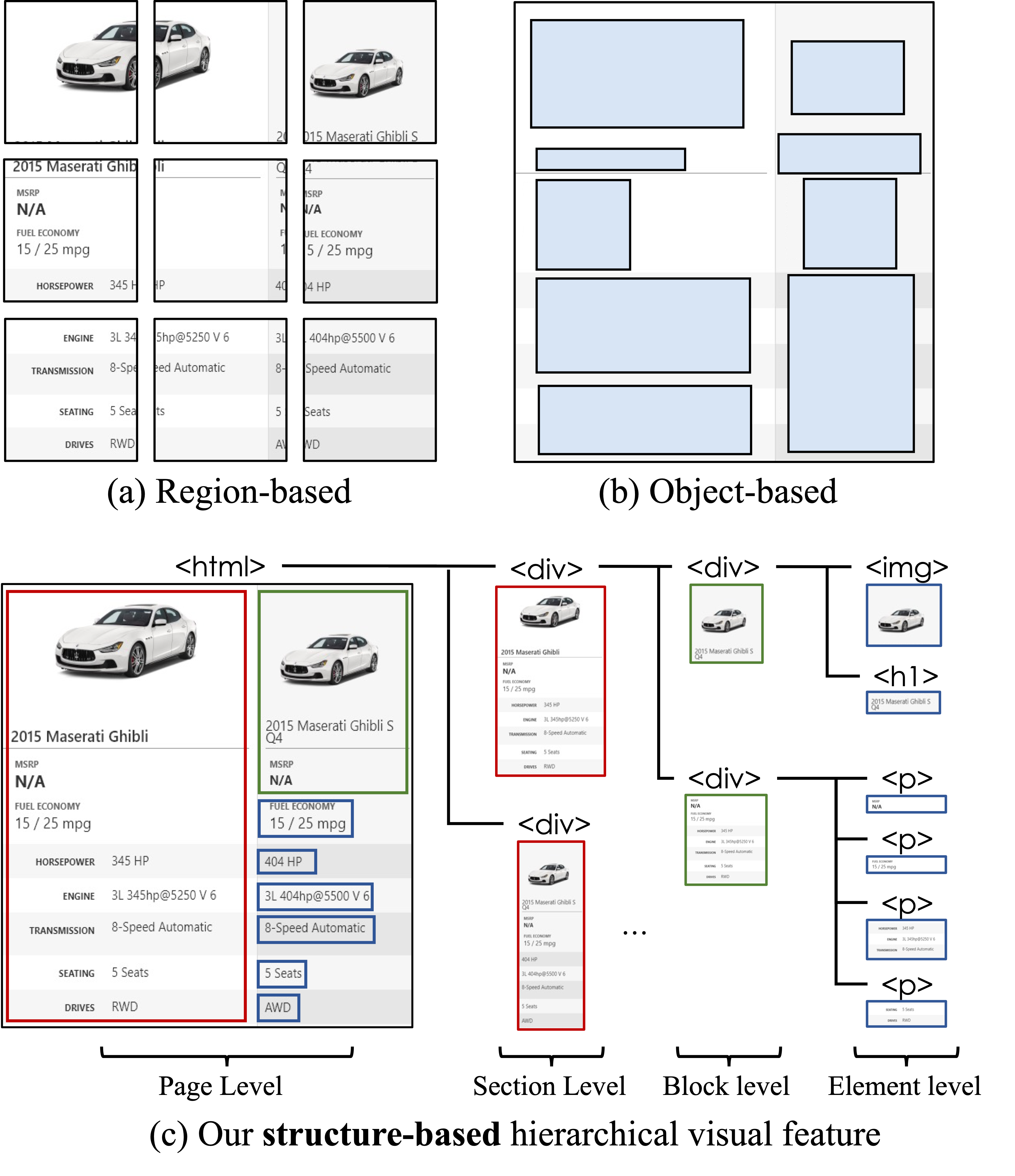}
    \caption{Comparison among different multimodal document pre-training methods.}
    \label{fig:feature_comparison}
\end{figure}
Multimodal pre-training models for documents can be broadly classified into two categories, namely image-oriented and text-oriented, based on the target document type. Image-oriented methods \cite{xu2020layoutlm,xu2020layoutlmv2,huang2022layoutlmv3,li2021selfdoc,gu2021unidoc,appalaraju2021docformer} deal with scanned/digital-born documents, where document images are easily accessible, and textual portions are typically acquired using external optical character recognition (OCR) tools. Text-oriented methods \cite{li2021markuplm,guo2022webformer,deng2022dom,gur2022understanding}, on the other hand, are concerned with markup-language-based documents such as webpages. Existing webpage pre-training models only use HTML as input, which consists of the structured description language and natural language content, i.e., the structure and text modalities.  This situation arises mainly because the current web pre-training datasets are either HTML-only like Common Crawl \footnote{https://commoncrawl.org/}, or failed to reach the pre-training scale.  However, it is impossible to understand real-world webpages using HTML alone due to the lack of information from other resources as we discussed in \cref{subsec:render}.


Furthermore, it is hard to seamlessly apply current multimodal pre-training methods to web pages. These approaches often treat document images as natural images, neglecting the structural complexities inherent in documents. As depicted in Figure~\ref{fig:feature_comparison}(a), \textbf{region-based} methods like LayoutLMv3\cite{huang2022layoutlmv3} partitions images into regions to extract region-level features, while \textbf{object-based} methods\cite{li2021selfdoc} in Figure~\ref{fig:feature_comparison}(b) rely on external tools to identify document objects and then extract object-level features. Unfortunately, both two types of methods neither capture multi-granularity visual features nor model the semantic relationship among those features. On the one hand, webpages exhibit a hierarchical structure, ranging from pages to sections, regions, and elements. External tools such as optical character recognition (OCR) or object detectors only recognize objects at a specific granularity, yet they are unable to capture features across various levels. On the other hand, there exist diverse semantic relationships between visual features of webpages, such as the sibling relationship between two elements or the parent-child relationship between the element and its parent section. The key to modeling such relationships is the structure of webpages that previous methods struggle to encode.

To address the above problems, we first collect a large-scale multimodal dataset for webpage pre-training, comprising a collection of \textbf{6 million} webpages from over \textbf{60,000} domains. This dataset encompasses HTML code, screenshots, and corresponding metadata. Second, we propose \textbf{WebLM}, a unified Transformer framework that concurrently models text, structure (markup language), and image modalities for understanding webpages. As shown in Figure~\ref{fig:feature_comparison}(c), WebLM is able to extract hierarchical visual features with the incorporation of HTML structure. This is implemented by considering the visual alignment between HTML tags and image regions contained in the metadata of our datasets. Last, we propose two novel pre-training tasks: \textbf{Tree Structure Prediction (TSP)}, focuses on predicting the tree-relationship between HTML nodes, which models the semantic structure of webpages both textually and visually;  \textbf{Visual Misalignment Detection(VMD)}, incorporates noise into the image region of HTML tags, compelling the model to be robust to the visual alignment between the two modalities.

We evaluate the WebLM models on the Web-based Structural Reading Comprehension (WebSRC) \cite{chen2021websrc} dataset and the Structured Web Data Extraction (SWDE) \cite{hao2011one}dataset. Experimental results show that our WebLM significantly outperforms previous SOTA pre-trained models. Ablation studies further demonstrate the effectiveness of incorporating the hierarchical visual feature. 

The contributions of this paper are summarized as follows:
\begin{itemize}
    \item We collect a large-scale multimodal dataset of \textbf{6 million} webpages from over \textbf{60,000} domains. The dataset, pre-trained models, and code are publicly available.
    \item We propose WebLM for webpage understanding, which introduces hierarchical visual features by first incorporating HTML structure into visual feature extraction. 
    \item We propose two novel pre-training tasks to effectively model the semantic structure of webpages and enhance the visual robustness of WebLM.
\end{itemize}

\section{Related Work}
\subsection{Multi-modal Document Pre-training}
Visually rich documents can be roughly divided into two categories based on the modalities involved: one is \textbf{image-centric} documents with image modality at the core, such as receipts and PDFs, where tasks often provide only image information and require external tools like OCR to obtain text and its location; the other is \textbf{text-centric} web documents, where the document image needs to be interactively and dynamically rendered based on the markup-language-based documents such as HTML/XML.

For scanned/digital-born documents, current pre-training methods often focus on extracting different granularity of visual features and then modeling the modality interaction via pre-training tasks. Both LayoutLM \cite{xu2020layoutlm} and LayoutLMv2 
\cite{xu2020layoutlmv2} uses ResNet-101 to extract fine-grained token-level visual features, whereas LayoutLMv2 introduces Text-Image Alignment and Text-Image Matching tasks to enhance the modality interaction. LayoutLMv3 \cite{huang2022layoutlmv3} and DiT \cite{li2022dit} adopt the encoding approach of Image Transformers (such as ViT \cite{dosovitskiy2020image}) for document image encoding. However, previous methods always overlook the hierarchical structure of documents, and the association between images and text is often provided by off-the-shelf tools, making it difficult for the model to learn.

For mark-language-based documents represented by webpages, existing pre-training models mainly focus on encoding the HTML source code, emphasizing the interaction between textual and structure modalities. MarkupLM \cite{li2021markuplm} inputs the text token sequence of HTML code and incorporates the xpath of each text's node as relation embedding. DOM-LM \cite{deng2022dom} extracts various structured information for text tokens, such as depth, tag type, and node index. Webformer \cite{guo2022webformer} designs a recursive encoding method for the tree structure of webpages. Furthermore, HTLM \cite{aghajanyan2021htlm} focuses on zero-shot prompting through HTML-based pre-training. Those pre-training models mainly utilize HTML as inputs while ignoring the image modality. However, as discussed in \cref{subsec:render}, HTML code in webpages contains only a portion of the information, while more style and structural information are found in webpage screenshots. 

\subsection{Webpages Understanding}
Webpages are the primary means for people to store, display, and transmit information on the Internet, making the automatic understanding and information extraction of webpages a blooming research topic. 
Hao et al. \cite{hao2011one} proposed the SWDE dataset for information extraction on webpages. Tanaka et al., as well as Chen et al., introduced web-oriented reading comprehension datasets VisualMRC \cite{tanaka2021visualmrc} and WebSRC \cite{chen2021websrc}, respectively, requiring models to understand the spatial structure of webpages as well as the textual content to answer corresponding questions. At the same time, many approaches \cite{qian2018graphie, lockard2020zeroshotceres, zhao2022tie} employ graph neural networks to encode node relationships in webpages. Additionally, large language models \cite{gur2022understanding} have been proven to possess strong webpage understanding capabilities via few-shot learning.

\section{WebLM}

WebLM is a multimodal pre-training framework that incorporates HTML structure into visual feature extraction. The motivation of WebLM is discussed in \cref{subsec:render}, followed by the introduction of model architecture (\cref{subsec:model_arch}) and pre-training tasks (\cref{subsec:pre_obj}).

\subsection{Motivation of WebLM}
\label{subsec:render}

\begin{figure}[t]
   \centering\includegraphics[width=\linewidth]{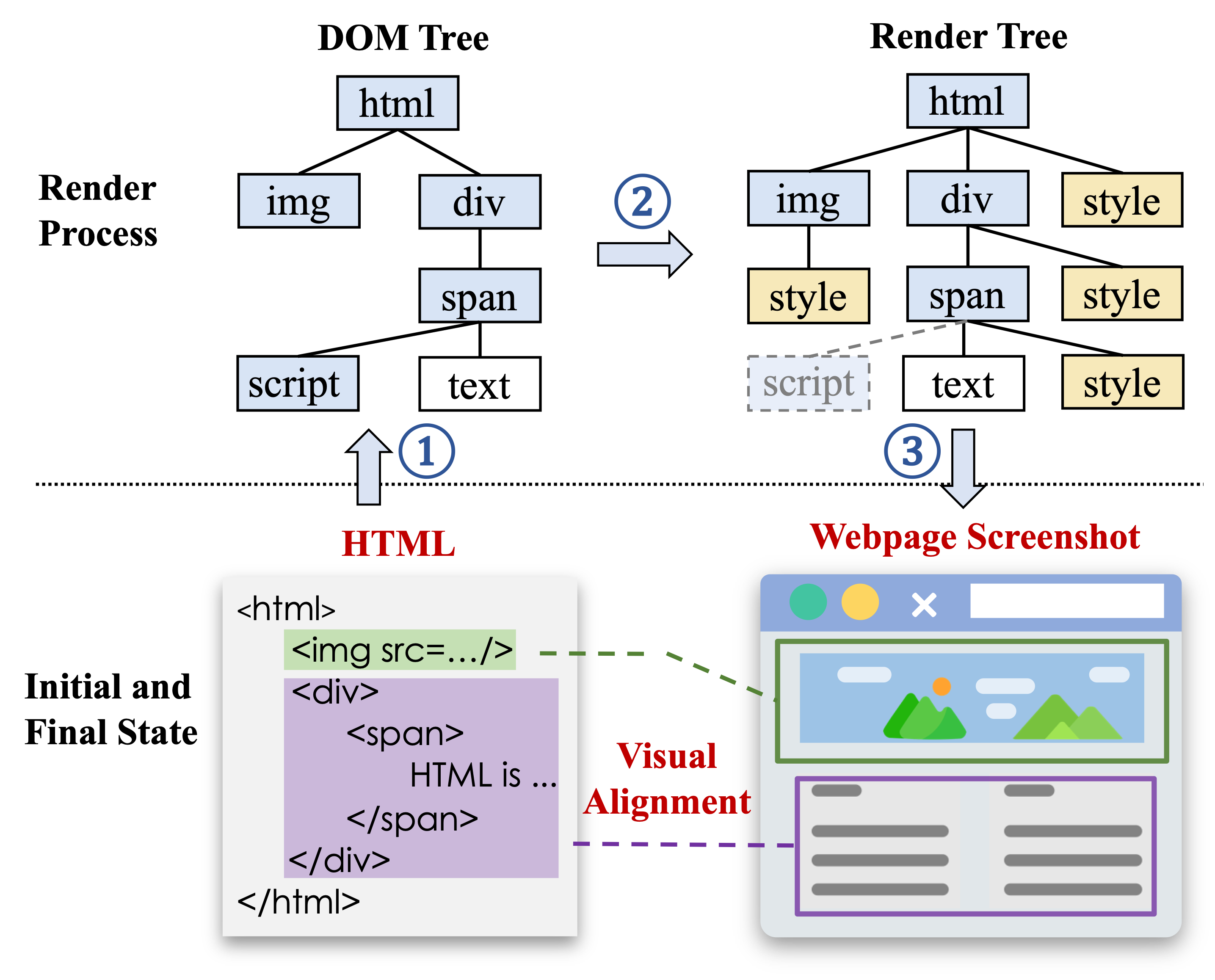}
    \caption{The rendering process from HTML to a webpage.}
    \label{fig:render}
\end{figure}

We show the overall rendering process of webpages in Figure \ref{fig:render}.
As we can see, it is impossible to represent and understand real-world webpages using HTML alone. This is mainly due to two reasons:
\begin{itemize}
    \item HTML does not contain information from external files, such as JavaScript, CSS, images, and other resources.
    \item Even with all the resources, rendering a webpage still requires browsers to interpret and execute the code, a complex process that existing models struggle to learn.
\end{itemize}

\begin{figure*}[t]
   \centering\includegraphics[width=0.95\textwidth]{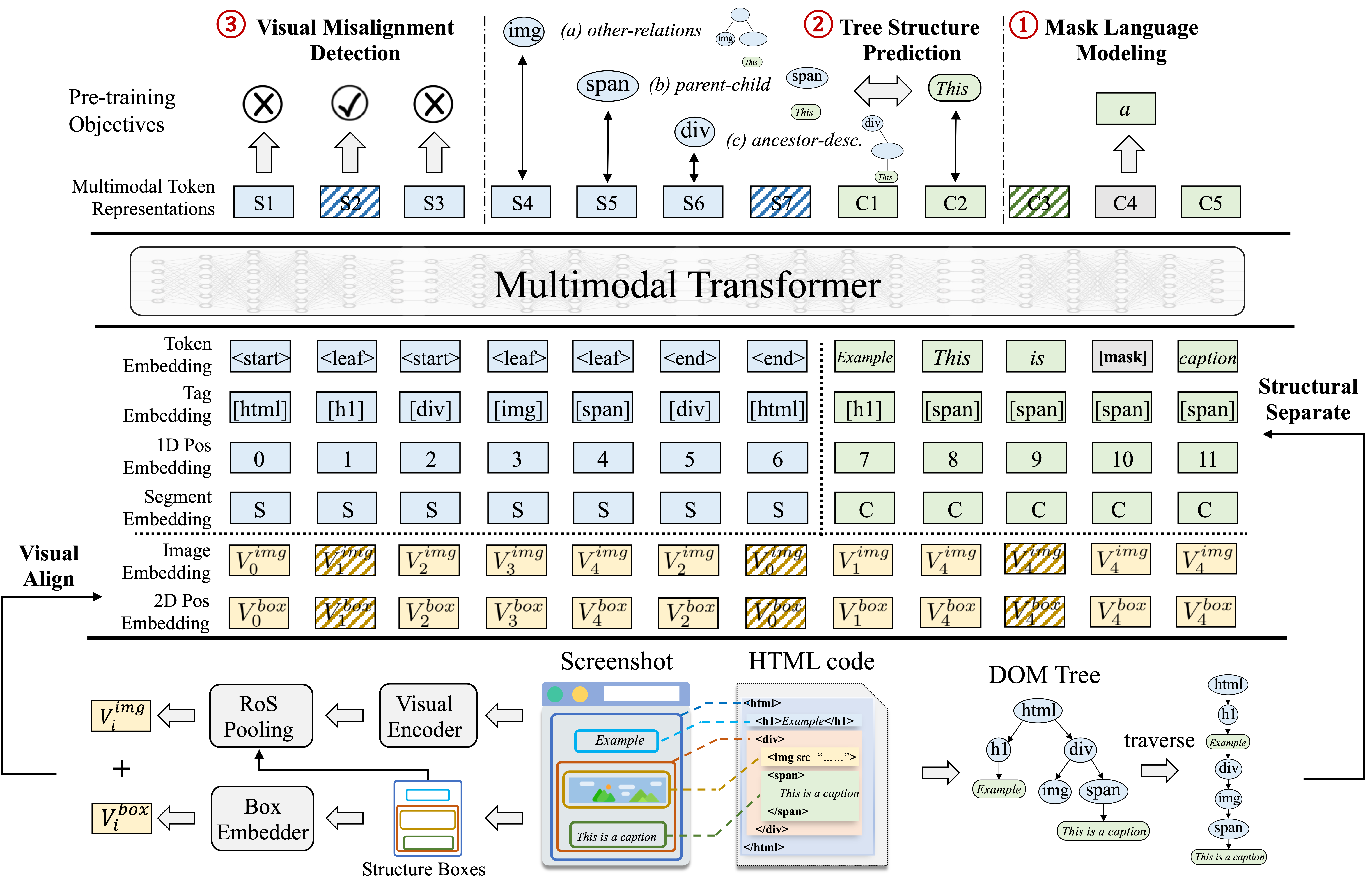}
    \caption{The architecture and pre-training objectives of WebLM.  The blue and green parts represent structure and content inputs from HTML code respectively, while the yellow part corresponds to visual inputs from webpage screenshots.}
    \label{fig:model_architecture}
\end{figure*}

Therefore, we believe that understanding webpages necessitates a multimodal approach, simultaneously incorporating both HTML code and webpage screenshots. On the one hand, screenshots contain the most complete style information, while HTML encompasses all content information, ensuring the coverage of all essential information. On the other hand, HTML represents the initial state of the webpage, and the screenshot corresponds to the final state. Employing these two states for webpage understanding eliminates the need to learn browser rendering logic. Moreover, the completion 
 of the rendering process results in a direct \textbf{visual alignment} between each node in the HTML and the region in the screenshot. WebLM leverages this alignment to provide visual features for each level of nodes in the HTML tree, effectively fusing structural and visual information to obtain hierarchical features of the images.

\subsection{Model Architecture}
\label{subsec:model_arch}

WebLM applies a unified multimodal Transformer to learn cross-modal 
representations where Figure \ref{fig:model_architecture} gives the architecture overview. The Transformer encoding layer is similar to BERT's \cite{devlin2018bert}, with key alterations made at the input layer. 
As shown in Figure~\ref{fig:model_architecture}, the input primarily comprises information from three modalities, corresponding to three different colors. Structure modality and content modality mainly come from HTML code, while visual modality comes from webpage screenshots. The primary design principles of the WebLM input layer are as follows: 
\begin{itemize}
    \item \textbf{Separate structure tokens from content tokens.} Previous models either directly take HTML code as input \cite{chen2021websrc, gur2022understanding}, or solely input the textual token of HTML while considering structure modality as supplementary features~\cite{li2021markuplm}. In contrast, WebLM takes both structure and textual tokens of HTML as input while separating the two modalities. This approach not only retains full document structure information by preserving the structure tokens but also accelerates information flow within each modality.
    \item \textbf{Align visual features with HTML inputs.} Most multimodal pre-training methods \cite{xu2020layoutlmv2,huang2022layoutlmv3,li2021selfdoc}regard the extracted sequence of visual features as a separate input sequence with respect to text modalities.  However, the alignment between text and image modalities is directly available in the context of webpages. Thus it is intuitive and reasonable to align visual features with text modalities before inputting them into the model rather than interacting during pre-training.
\end{itemize}

\noindent We further introduce the embedding details of each modality:

\paragraph{Token Embedding} To construct the input token sequence for the model, we first convert the HTML code into a DOM tree and traverse it in a depth-first order. During traversing, we perform \textbf{structural separate}, which places structure tokens and text tokens in two lists. After concatenating the two sequences, we get the final input token sequence like  
$$ {\rm T} = \{ {\tt[CLS]},  s_1, s_2, ..., {\tt[SEP]} c_1, c_2, ...,  {\tt[SEP]}\}.$$

It is worth noting that our structure tokens only include the HTML tags themselves, excluding their attributes, as attribute information introduces significant noise. Furthermore, we simplify all tags into three types: \texttt{<start\_tag>}, \texttt{<end\_tag>}, and \texttt{<leaf\_tag>}, and add tag types as additional embeddings. In fact, HTML tags have two main functions: one is to express the tree structure of HTML  through the correspondence between pairs of opening and closing tags (e.g., \texttt{<p>} and \texttt{</p>}) and their nesting; the other is to indicate tag-specific roles, such as <h1> for headings and <img> for images.
With approximately 120 common tags and extremely imbalanced distribution (e.g., the frequency of \texttt{<div>} tag is much higher than that of other tags. ), separating structural and functional embeddings of tags allows for shared learning for the structural embedding of each tag, thus better modeling the structure of HTML.

\paragraph{Tag Embedding}
There are approximately 120 common HTML tags, which can be classified according to their functions. 
Understanding the function of each tag helps to better interpret web content. Furthermore, each common tag has a textual description corresponding to its function, which can effectively aid the model in understanding the tag's purpose. Thus We employ 
\texttt{sentence- Transformers} \footnote{https://www.sbert.net/} to extract an embedding vector from each tag's textual description \footnote{https://www.w3schools.com/tags/default.asp} as the tag's initial embedding. For infrequent tags, we convert them to \texttt{<unk>} and initialize it randomly. 

The final \textbf{text embedding} from HTML is the sum of four embeddings. Token embedding and 1D positional embedding 
represents the token and its index. 
Tag embedding represents the function of the token's corresponding HTML tag.
For each token $w_i$ from either structure input or content input, we incorporate the tag embedding based on the tag type of their respective nodes ${tag}_i \in \{\texttt{<html>}, \texttt{body}, \texttt{<div>}, ...\}$in the DOM tree.  Besides, we use segment embedding to distinguish structure and text content tokens by assigning each token to a segment ${seg}_i \in \{{\rm \texttt{[S]}}, {\rm \texttt{[C]}}\} $,  Formally, we have the $i$-th text embedding for a token $w_i$:
$${\rm \textbf{t}}_i = {\rm TokEmb}(w_i) + {\rm TagEmb}({tag}_i) + {\rm PosEmb1D}(i) + {\rm SegEmb}({seg}_i).$$

\paragraph{Image Embedding} WebLM employs a ResNeXt-FPN \cite{xie2017aggregated, lin2017feature} architecture as the backbone of the visual encoder. Given a webpage screenshot ${\rm I}$, it is resized to 224 × 224 and fed into the visual backbone. Then WebLM  extracts the corresponding visual feature for all the nodes on the HTML DOM tree. Specifically, while object detection models perform pooling on RoI (Regions of Interest) \cite{he2016deep}, WebLM conducts pooling on \textbf{RoS} (Regions of Structure nodes on the DOM tree) according to the visual alignment information of each node. After obtaining the image embeddings for each node on the DOM tree, we perform \textbf{visual align} by aligning the node visual feature sequence with the input token sequence based on the correspondence between token and node, allowing each token to obtain its associated visual features. The image embedding of $i$-th token with the corresponding node $n_i$ is formulated as
$${\rm v}^{img}_{i} = {\rm RoSPool}({\rm VisualEncoder(I)})_{n_i}$$

\paragraph{2D Position Embedding} The image feature extracted by the visual encoder primarily contains style information, such as color, and font. In addition, we input the bounding box of the node region to embed spatial layout information. Following previous works \cite{xu2020layoutlmv2}, we normalize and discretize all coordinates to integers in the range $[0, 1000]$, and use two embedding layers to embed x-axis features and y-axis features separately. Given the $i$-th token and the normalized bounding box of its corresponding node $n_i$ ${\rm box}_{n_i} = (x_0, x_1, y_0, y_1, w, h)$, we calculate the 2D position embedding by concatenating six bounding box features and aligning it to $i$-th token:
$${\rm v}^{box}_{i} = {\rm Concat}({\rm PosEmb2D_x}(x_0, x_1, w), {\rm PosEmb2D_y}(y_0, y_1, h))$$

The final \textbf{visual embedding} of the $i$-th token is the sum of its image embedding and 2D position embedding
$${\rm v}_i = {\rm v}^{img}_{i} + {\rm v}^{box}_{i},$$
and we obtain the final input embedding ${\rm x}_i$ of the $i$-th token by adding its text embedding ${\rm t}_i$ and visual embedding ${\rm v}_i.$

\subsection{Pre-training Objectives}
To efficiently model the complex structure of webpages and enhance the information exchange among the three modalities, We design three self-supervised pre-training tasks for WebLM, including mixed-modality MLM and cross-modality TSP and VMD tasks.
\label{subsec:pre_obj}

\textbf{Masked Language Modeling (MLM).}
Following previous works \cite{devlin2018bert, xu2020layoutlm}, we use MLM to enhance the model's language understanding capabilities. We randomly replace some content tokens with \texttt{[MASK]} and require the model to predict the original words. Unlike the Masked Visual-Language Modeling in LayoutLMv2 \cite{xu2020layoutlmv2}, we do not mask the corresponding regions of tokens in the image due to the unavailability of token-level region positions.

\textbf{Tree Structure Prediction (TSP).}
We propose the TSP task based on the following two observations:
\begin{itemize}
    \item While the structural separation facilitates intra-modality information flow, it slows down cross-modality information flow between structure and content modalities.
    \item The tree structure of HTML explicitly conveys the main semantic structure of both textual and visual inputs.
\end{itemize}
Thus the TSP task requires WebLM to predict the tree relationship between structure and content inputs to accelerate cross-modality information flow as well as modeling the semantic structure of webpages. Specifically, we sample a node token from the structural input and a text token from the content input, requiring the model to determine their relationship based on the DOM Tree $R \in \{ $ \texttt{parent-child}, \texttt{ancestor-descendent}, \texttt{other-relations}$\}$. In addition, text tokens within the tree solely constitute leaf nodes, yet their quantity far surpasses that of structural nodes. Structural separation helps TSP to sample more diverse tree node pairs while simplifying the implementation via complex sampling algorithms.

\textbf{Visual Misalignment Detection (VMD).} As we described in sec~\ref{subsec:render}, visual alignment between HTML nodes and screenshot regions is the key to constructing the multimodal input sequence of WebLM. 
Due to rendering issues or external interference, such alignment might introduce noise, potentially impacting the model's performance. Therefore, the TIM task is proposed to enhance the visual robustness.
Specifically, we randomly sample some positions from the input sequence (including both structural and content inputs) and add noise to their corresponding visual feature regions, either enlarging or reducing them by 50\%. The model is then asked to identify if the image information at each token has been affected by noise. This perturbing simultaneously alters both the 2D position embedding and image embedding, requiring the model to make judgments based on the textual modality or the semantic relationships between with surrounding tokens.

\begin{table*}[t]
    \caption{Evaluation results on  WebSRC. EM, F1, POS denotes the exact match score, the token level F1 score, and the path overlap score, respectively. We submit the models to the official of WebSRC for testing. * denotes reproduction results.}
    \centering
    \begin{tabular}{p{0.015\textwidth}<{\centering}|p{0.25\textwidth}
    <{\centering}|p{0.15\textwidth}
    <{\centering}|p{0.05\textwidth}<{\centering}p{0.05\textwidth}<{\centering}p{0.05\textwidth}<{\centering}|p{0.05\textwidth}<{\centering}p{0.05\textwidth}<{\centering}p{0.05\textwidth}<{\centering}}
         \hline

        \hline
         
         &\multirow{2}{*}{Method} & \multirow{2}{*}{Modalities} & \multicolumn{3}{c|}{Dev} & \multicolumn{3}{c}{Test} \\
         \cline{4-9}
         &&& EM$\uparrow$ & F1$\uparrow$ & POS$\uparrow$ & EM$\uparrow$ & F1$\uparrow$ & POS$\uparrow$ \\

                                         
         \midrule
             \multirow{8}{*}{\begin{sideways}BASE\end{sideways}} & T-PLM(BERT)~\citep{chen2021websrc} & \multicolumn{1}{l|}{Text} & 52.12 & 61.57 & 79.74 & 39.28 & 49.49 & 67.68 \\
         &H-PLM(BERT)~\citep{chen2021websrc} & \multicolumn{1}{l|}{Text + HTML}  & 61.51 & 67.04 & 82.97 & 52.61 & 59.88 & 76.13 \\
         &V-PLM(BERT)~\citep{chen2021websrc} & \multicolumn{1}{l|}{Text + HTML + Image} & 62.07 & 66.66 & 83.64 & 52.84 & 60.80 & 76.39 \\
         \cdashline{2-9}

         &$\text{DOM-LM}$~\citep{deng2022dom} & \multicolumn{1}{l|}{Text + HTML} & 69.70 & 73.90 & - & - & - & - \\
         & $\text{LayoutLMv3}^*$~\citep{huang2022layoutlmv3} & \multicolumn{1}{l|}{Text + Image} & 66.33 & 71.46 & 85.27 & 48.33 & 51.64 & 71.02 \\
         &$\text{MarkupLM}$~\citep{li2021markuplm}& \multicolumn{1}{l|}{Text + HTML} & 68.39 & 74.47 & 87.93 & - & - & - 
         \\
          & $\text{MarkupLM}^{*}$ & \multicolumn{1}{l|}{Text + HTML} & 68.99 & 74.55 & 88.40 & 60.43 & 67.05 & 80.55 \\
         \cline{2-9}
          \rowcolor{blue!5} & $\textbf{WebLM}$ & \multicolumn{1}{l|}{Text + HTML + Image} & \textbf{72.14} & \textbf{79.67} & \textbf{89.36} & \textbf{65.95} & \textbf{72.30} & \textbf{83.77} \\
         \midrule
         \multirow{7}{*}{\begin{sideways}LARGE\end{sideways}} & T-PLM(Electra)~\citep{chen2021websrc} & \multicolumn{1}{l|}{Text} & 61.67 & 69.85 & 84.15 & 56.32 & 72.35 & 79.18 \\
         &H-PLM(Electra)~\citep{chen2021websrc}& \multicolumn{1}{l|}{Text + HTML}  & 70.12 & 74.14 & 86.33 & 66.29 & 72.71 & 83.17 \\
         &V-PLM(Electra)~\citep{chen2021websrc}& \multicolumn{1}{l|}{Text + HTML + Image}  & 73.22 & 76.16 & 87.06 & 68.07 & 75.25 & 84.96 \\

         \cdashline{2-9}

         & $\text{LayoutLMv3}^*$~\citep{huang2022layoutlmv3}& \multicolumn{1}{l|}{Text + Image} & 71.38 & 75.73 & 87.74 & 57.68 & 63.33 & 79.76 \\
         
         &$\text{MarkupLM}$~\citep{li2021markuplm}& \multicolumn{1}{l|}{Text + HTML}  & 74.43 & 80.54 & 90.15 & - & - & - \\
         &$\text{MarkupLM}^*$& \multicolumn{1}{l|}{Text + HTML} & 73.38 & 79.83 & 89.93 & 69.09 & 76.45 & 87.24 \\
         \cline{2-9}

         \rowcolor{blue!5} &$\textbf{WebLM}$ & \multicolumn{1}{l|}{Text + HTML + Image}  & \textbf{78.40} & \textbf{84.24} & \textbf{91.54} & \textbf{72.01} & \textbf{78.66} & \textbf{88.33} \\
         \hline

         \hline
    \end{tabular}
    \label{tab:websrc_table}
\end{table*}

\section{Experiments}

WebLM focuses on webpage understanding through multimodal pre-training. Therefore, we evaluate it on web-based question answering and information extraction tasks and further investigate the importance of each component through ablation studies.

\subsection{Data}
\subsubsection{Pre-training Data}

\paragraph{Common Crawl} Common Crawl is a publicly available web crawl dataset that collects webpages from the internet. Instead of using the source code, we collect webpage links from Common Crawl. We traverse all links in a dataset snapshot \footnote{https://commoncrawl.org/2022/08/august-2022-crawl-archive-now-available/} and categorize and sort them based on the domain name. We select 60,000 domains with the largest number of webpages, with each domain containing 100 webpages.  We also use \texttt{fasttext} \cite{joulin2016bag} to filter out non-English pages with an English classification score < 0.6. We then employ \texttt{Selenium} \footnote{https://www.selenium.dev/} to crawl the corresponding HTML, webpage screenshots, and bounding box information for each HTML node. Finally, we obtain a dataset of 6 million webpages for WebLM pre-training. 

\paragraph{Pre-processing} 
Due to a large number of textual tokens of real-world webpages and the typically long screenshots, it is challenging to input entire webpages into the model for pre-training. Consequently, we construct pre-training features mainly through two approaches: simplifying the HTML input and extracting input segments from complete webpages. We only retain nodes on the HTML rendering tree that either contain text or have corresponding regions in the image, while removing nodes that do not have either (e.g., \texttt{<script>}, \texttt{<style>}, etc.). Furthermore, we simplify the HTML structure by replacing a node with its child node if it has only one child. After these modifications, we traverse the HTML tree and search for nodes whose total number of tokens in their child nodes and text lies within a certain range (i.e., between 128 and 512.) as input features. By extracting the corresponding screenshot area and HTML code segment, we construct features that meet input length constraints. Additionally, if combined sibling nodes satisfy the input length limits, we also use their corresponding HTML code segments and largest bounding box screenshots as input features.

\subsubsection{Fine-tuning Data}

\paragraph{WebSRC} 
WebSRC \cite{chen2021websrc} is a Web-based Structural Reading Comprehension dataset that aims to test the ability of models to understand the contents of webpages as well as their structures.  WebSRC consists of 400K question-answer pairs, which are collected from 6.4K webpages. 
Each question in WebSRC requires a certain structural understanding of a webpage to answer, and the answer is either a text span on the webpage or yes/no. Following the original paper, we use \textbf{Exact match (EM)}, \textbf{F1 score (F1)}, and \textbf{Path overlap score (POS)} as evaluation metrics.

\paragraph{SWDE}
The Structured Web Data Extraction (SWDE) \cite{hao2011one} dataset is a real-world collection of webpages used for automatic information extraction. It consists of 8 verticals, 80 websites (10 per vertical), and 124,291 webpages in total. The goal is to extract values corresponding to given attributes from a webpage, such as the \textit{price} value in \textit{shopping} pages. We use \textbf{page-level F1} scores as our evaluation metric as in previous works \cite{zhou2021simplified, lin2020freedom, li2021markuplm}. 
We follow MarkupLM to train and evaluate each vertical independently. In each vertical, we select $k$ consecutive seed websites for training and use the remaining $10-k$ websites for testing. 
The final results are obtained by averaging across all 8 verticals and all 10 permutations of seed websites per vertical, resulting in 80 experiments for each $k$.

\subsection{Experiment Setup}
\label{sec:settings}
\paragraph{Pre-training}
The token-masked probability in MLM and visually noise-adding probability in VMD are both 15\%. The probability of the bounding box increasing or decreasing in size is each 50\%. The max number of selected node pairs is 1,000 in TSP for each sample, and we limit the ratio of pairs with \texttt{other-relations} as 60\% to make a balance. We initialize $\textrm{WebLM}$ from $\textrm{RoBERTa}$ and train the base and large model for 300K steps on 8 NVIDIA A10 and A100 GPUs, respectively. For the ResNeXt-FPN part in the visual embedding layer, the backbone of a Mask-RCNN \cite{he2017mask} model trained on PubLayNet \cite{zhong2019publaynet} is leveraged \footnote{“MaskRCNN ResNeXt101 32x8d FPN 3X” setting in https://github.com/hpanwar08 /detectron2}. We set the total batch size as 256, the learning rate as 5e-5, the max sequence length as 512, and the warmup ratio as 0.1. The selected optimizer is AdamW \cite{loshchilov2017decoupled}, with $\epsilon=1e-6$, $\beta_1=0.9$, $\beta_2=0.98$, $\texttt{weight decay}=0.01$, and a linear decay learning rate scheduler with 6\% warmup steps. We also apply $\texttt{FP16}$ to reduce GPU memory consumption and accelerate training.

\paragraph{Fine-tuning}
We treat the WebSRC task and SWDE task as an extractive QA task and token classification task, respectively. In the input layer, we truncate any structural tokens exceeding a fixed length and concatenate the HTML text and the question as the content input. When the content length
surpasses the limit, a sliding window mechanism is employed for
multiple inputs. For separators such as [CLS] and [SEP], as well as
question tokens, we consider them to be directly connected to the
<html> node. For WebSRC, we fine-tune WebLM for 2 epochs with a total batch size of 64 and a learning rate of 1e-5. For SWDE, we fine-tune WebLM with 10 epochs, a total batch size of 64, and a learning rate of 2e-5. The warmup ratio is set to 0.1 and the max sequence length is set as 512 in both tasks, and we keep other hyper-parameters as default.

\subsection{Baselines}
We only introduce the SOTA pretrained models here and refer readers to \cite{chen2021websrc,zhou2021simplified} for more details above non-pretrained baselines:

\noindent\textbf{DOM-LM.} DOM-LM\cite{deng2022dom} is an HTML-based pre-trained model that takes text tokens and several HTML DOM tree features as inputs, such as depth, tag type, and node index. 

\noindent\textbf{LayoutLMv3.} LayoutLMv3\cite{huang2022layoutlmv3} is a multimodal pre-trained model for document understanding.  It simplifies LayoutLMv2\cite{xu2020layoutlmv2} by using patch embeddings (as in ViT) instead of leveraging a CNN backbone. LayoutLMv3 exhibits a general capacity for visual understanding while having a modest performance on textual modeling.

\noindent\textbf{MarkupLM.} MarkupLM\cite{li2021markuplm} is a SOTA webpage pre-trained model which only inputs the text token sequence of HTML code and incorporates the xpath of each text's node as supplementary information. Instead of explicitly modeling the structure of HTML, it regards the tree relationship of node pairs as a type of relation embedding between text tokens. Consequently, MarkupLM achieves the best text-understanding abilities among all models.



\subsection{Main Results}

\begin{figure}[t]
   \centering\includegraphics[width=\linewidth]{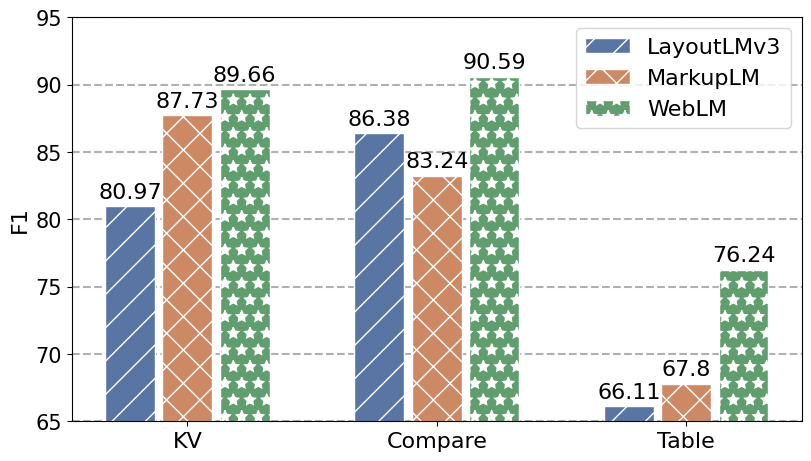}
    \caption{The performance comparison on different types of websites of WebSRC development set.}
    \label{fig:websrc_type_comparision}
\end{figure}

\begin{table}[t]
       \caption{Results on SWDE using different numbers of seed sites $k=\{1,2,3,4,5\}$. The baseline results are from \citep{zhou2021simplified}.}
    \centering
    \begin{tabular}{cccccc}
    \toprule
        Model $\backslash$ \#Seed Sites & $k=1$ &  $k=2$ &  $k=3$ &  $k=4$ &  $k=5$\\
    \midrule
        \texttt{SSM} \citep{carlson2008bootstrapping} & 63.00 & 64.50 & 69.20 & 71.90 & 74.10\\
        \texttt{Render-Full} \citep{hao2011one} & 84.30 & 86.00 & 86.80 & 88.40 & 88.60 \\
        \texttt{FreeDOM-NL} \citep{lin2020freedom} & 72.52 & 81.33 & 86.44 & 88.55 & 90.28 \\
        \texttt{FreeDOM-Full} \citep{lin2020freedom} & 82.32 & 86.36 & 90.49 & 91.29 & 92.56 \\
        \texttt{SimpDOM} \citep{zhou2021simplified} & 83.06 & 88.96 & 91.63 & 92.84 & 93.75\\
	\midrule
    	$\textrm{MarkupLM}_{\rm BASE}$ & 82.11& 91.29 & 94.42 & 95.31 & 95.89 \\
        $\textbf{WebLM}_{\textbf{BASE}}$ & \textbf{84.21} & \textbf{93.17} & \textbf{95.68} & \textbf{96.17} & \textbf{96.78} \\
        \cdashline{1-6}
        
    	$\textrm{MarkupLM}_{\rm LARGE}$ & 85.71 & 93.57 & 96.12 & 96.71 & 97.37\\
         $\textbf{WebLM}_{\textbf{LARGE}}$ & \textbf{87.57}& \textbf{94.89} & \textbf{97.25} & \textbf{97.54} & \textbf{98.10} \\
    \bottomrule 
    \end{tabular}
    \label{tab:swde}
\end{table}

As shown in Table \ref{tab:websrc_table}, both base and large versions of our proposed WebLM significantly outperform all baseline models on WebSRC dataset. Compared to MarkupLM, WebLM still exhibits substantial performance improvements. This demonstrates that WebLM can effectively utilize the information from all three modalities, achieving a better understanding of both webpage structure and textual content. Additionally, although LayoutLMv3 is not pre-trained on web data, it still exhibits good performance on the dev set, highlighting the importance of visual modality. However, its lower performance on the test set indicates a weaker generalization ability, emphasizing the necessity of pre-training on web data. 

We further compare the model performance on different types of websites as shown in Figure \ref{fig:websrc_type_comparision}. \texttt{KV}-type websites emphasize the comprehension of textual semantics, whereas \texttt{Compare} and \texttt{Table}-type websites underscore the significance of webpage sematic structure. We find that WebLM demonstrates a dual proficiency encompassing strong textual comprehension and better webpage structure modeling. Especially in visually complex webpages, i.e., \texttt{Compare} and \texttt{Table}-type websites, WebLM significantly outperforms the other two models. This success further demonstrates the necessity and effectiveness of introducing the hierarchical visual feature.

The results for the SWDE dataset are shown in Table \ref{tab:swde}. Since the SWDE dataset was created earlier, many webpages in the dataset do not contain visual information such as CSS and screenshots. Therefore, we render the HTML files in a browser to generate corresponding screenshots, which introduces a considerable amount of noise. Nevertheless, from the experimental results, we can observe that both the base and large models of our WebLM outperform MarkupLM. The performance improvement is more pronounced when the training data is limited, i.e., when k is small. This demonstrates that WebLM possesses robustness to visual information noise, maintaining a good webpage understanding capability even in noisy environments.

\subsection{Ablation Study}

\subsubsection{Pre-training Tasks}

\begin{table}[t]
    \caption{Ablation study of pre-training tasks on WebSRC dev set.}
  \centering
  \resizebox{\columnwidth}{!}{%
    \begin{tabular}{ccccccc}
    \toprule
         Pre-training Data & \multicolumn{3}{c}{Objectives} & \multicolumn{3}{c}{Metrics} \\
          \midrule
          Samples   & MLM   & TSP   & TIM   & EM & F1    & POS \\
    \midrule
    1M     & \checkmark & & &  64.17	& 72.13	& 86.33  \\
    1M     & \checkmark & \checkmark      &  & 66.99 &	74.92 &	87.78 \\
    1M     & \checkmark & \checkmark & \checkmark & 67.43      &  76.93    & 88.60 \\
    \midrule
    6M     & \checkmark & \checkmark & \checkmark & 72.14      & 79.67      & 89.36 \\
    \bottomrule

    \end{tabular}%
    }
  \label{tab:task_ablation}
\end{table}

The ablation results of pre-training tasks are shown in Table \ref{tab:task_ablation}. We found that both TSP and VMD, two pre-training tasks focusing on inter-modal interactions, significantly contribute to the model's performance. When removing the TSP task, which focuses on the interaction between structure and content modalities, WebLM's performance loss is greater, with a decrease of 2.8 points in both the EM score and F1 score, demonstrating that modeling HTML structure can better help the model understand web content. The VMD task enhances model performance by strengthening the interaction between text and image modalities. Furthermore, our WebLM pre-trained with just 1 million webpages performs on par with the MarkupLM model trained with 24 million webpages, which demonstrates the efficient utilization of our high-quality datasets.

\begin{figure}[t]
    \centering
    \includegraphics[width=0.95\linewidth]{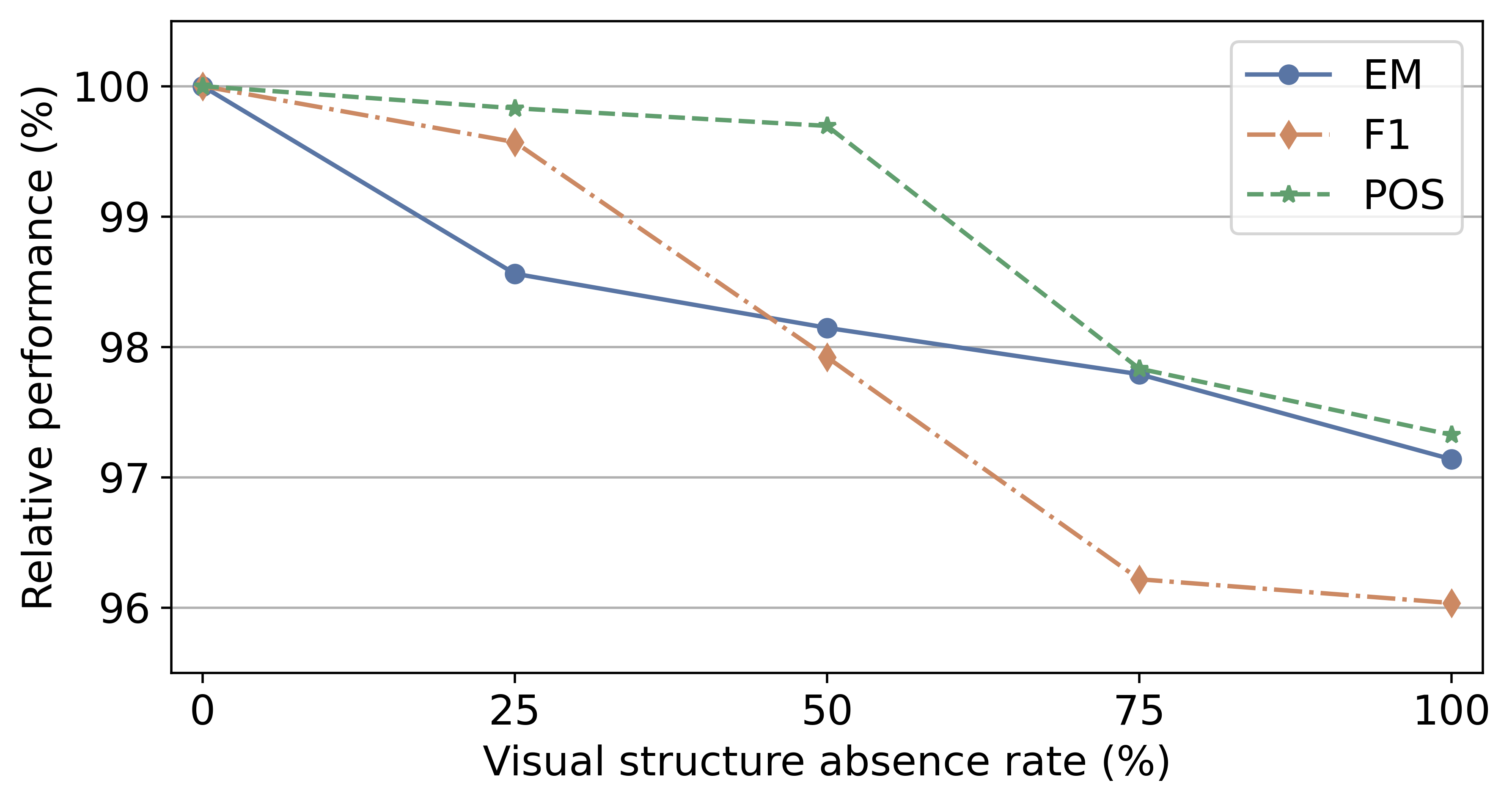}
    \caption{Relative performance on WebSRC Dev set compared to full-structured methods after truncating HTML structures.}
    \label{fig:structure_ablation}
\end{figure}

\subsubsection{Visual Features}

\begin{table}[t]
    \caption{Ablation study of visual embeddings on the WebSRC dev set.}
    \centering
    \resizebox{\columnwidth}{!}{\begin{tabular}{p{0.3cm}<{\centering}@{-}lccc}
         \toprule
         \multicolumn{2}{c}{Method} & EM$\uparrow$ & F1$\uparrow$ & POS$\uparrow$ \\
        \midrule

         \multicolumn{2}{l}{$\text{WebLM}_{\text{BASE}}$+MLM\&TSP} & 66.99 & 74.92 & 87.78 \\
         & w/o $\text{Image Embedding}$ & $\text{60.56}_{\text{(-6.43)}}$ & $\text{71.77}_{\text{(-3.15)}}$ & $\text{	85.81}_{\text{(-1.97)}}$ \\
         & w/o 2D Position Embedding& $\text{62.91}_{\text{(-4.08)}}$ & $\text{70.18}_{\textbf{(-4.74)}}$ & $\text{84.22}_{\textbf{(-3.56)}}$\\
         & w/o Visual Embedding & $\text{58.52}_{\textbf{(-8.47)}}$ & $\text{71.13}_{\text{(-3.79)}}$ & $\text{84.54}_{\text{(-3.24)}}$ \\
         \bottomrule
    \end{tabular}}
    \label{tab:visual_ablation}
\end{table}
The ablation study results of different visual embeddings are shown in Table \ref{tab:visual_ablation}. We find that both Image embedding and 2D position embedding have a strong impact on the model's performance. When these two features are removed, the model's performance experiences a sharp decline. Moreover, Image embedding has a greater influence on predicting the EM score, while 2D position embedding has a more significant impact on predicting the F1 score. When both features are removed, the task's EM score experiences a more substantial decrease. This demonstrates the importance of visual features and further confirms WebLM's ability to effectively incorporate and utilize visual features.

\subsubsection{Hierarchically Visual Structure.}

While previous work can also obtain features of each sub-region within an image, our method uniquely leverages the HTML structure to hierarchically combine these sub-region features. This generation of visual features is an attribute absent in prior studies. We also demonstrate its effectiveness with additional experiments. As shown in Figure~\ref{fig:structure_ablation}, we truncate HTML non-leaf nodes that are below a specific depth (-25\% signifies truncating those nodes that are less than 25\% of the maximum depth). This approach allows for the preservation of all fine-grained visual information in the image while eliminating hierarchical visual features of other granularities. Our experiments demonstrate that even with the inclusion of visual information from images, there is a substantial decline in model performance if the incorporation of HTML structure is omitted.

\begin{figure}[t]
    \centering
    \begin{subfigure}[t]{0.32\linewidth}
           \centering
           \includegraphics[width=\linewidth]{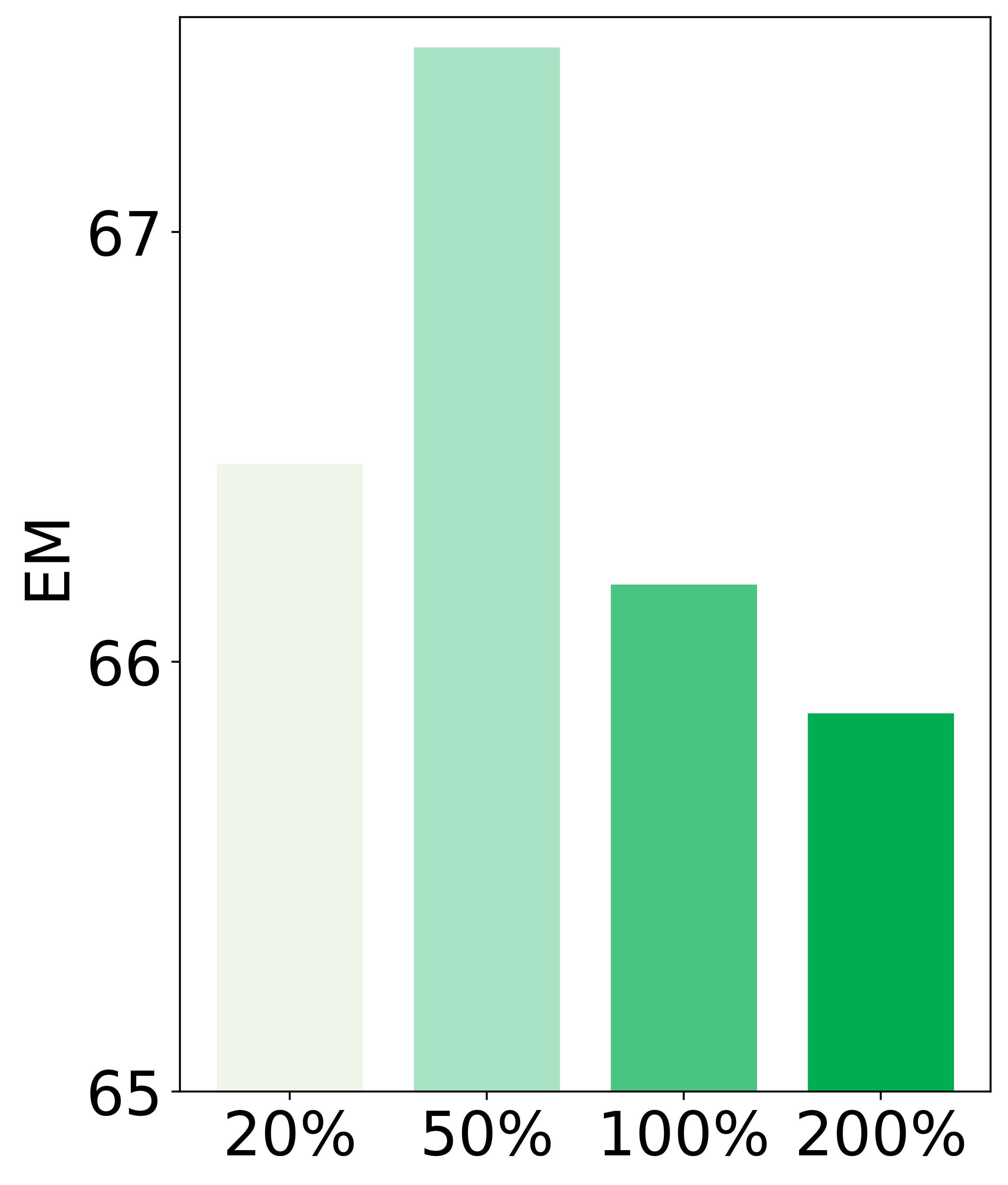}
    \end{subfigure}
    \begin{subfigure}[t]{0.32\linewidth}
            \centering
            \includegraphics[width=\linewidth]{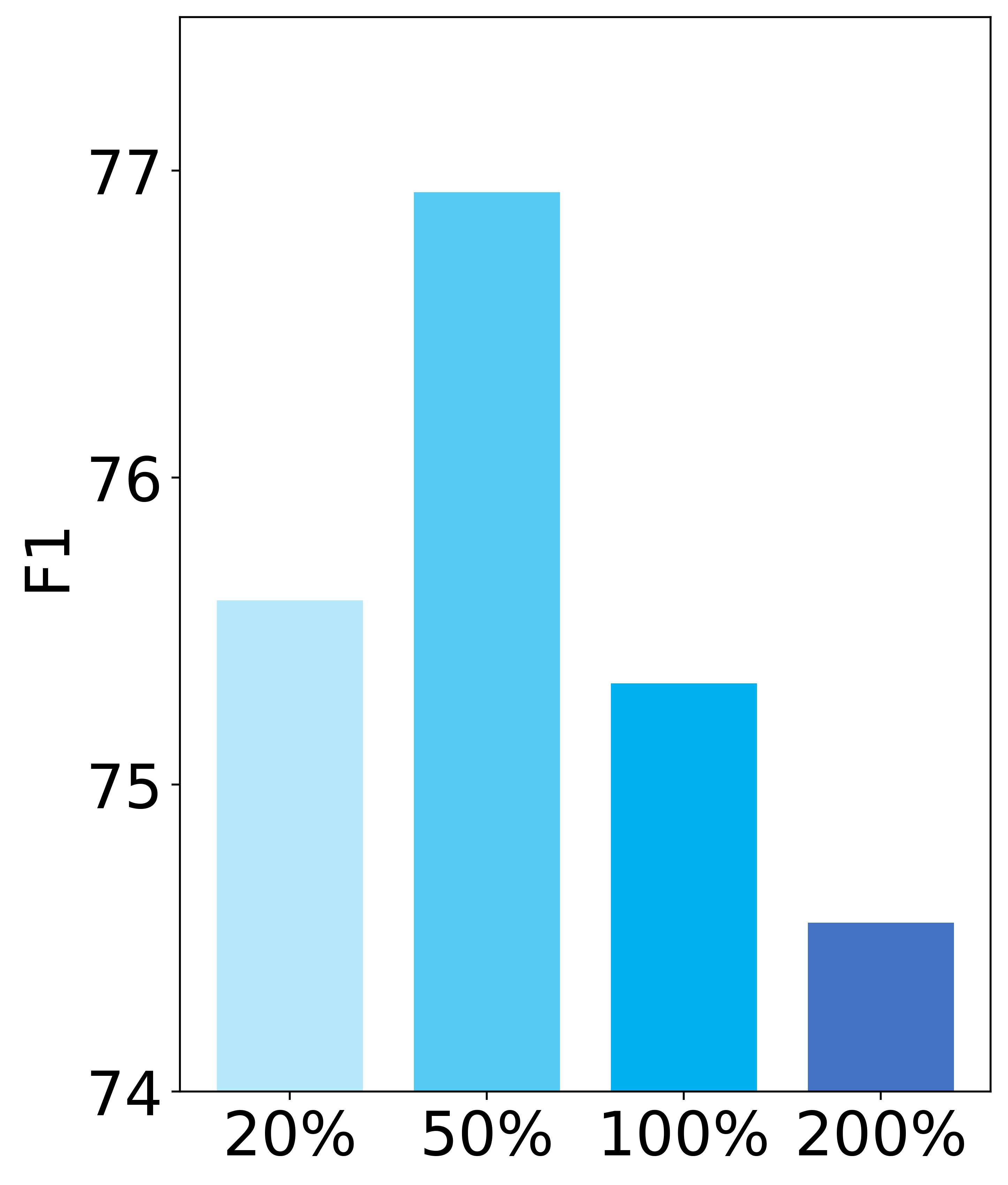}
    \end{subfigure}
    \begin{subfigure}[t]{0.32\linewidth}
            \centering
            \includegraphics[width=\linewidth]{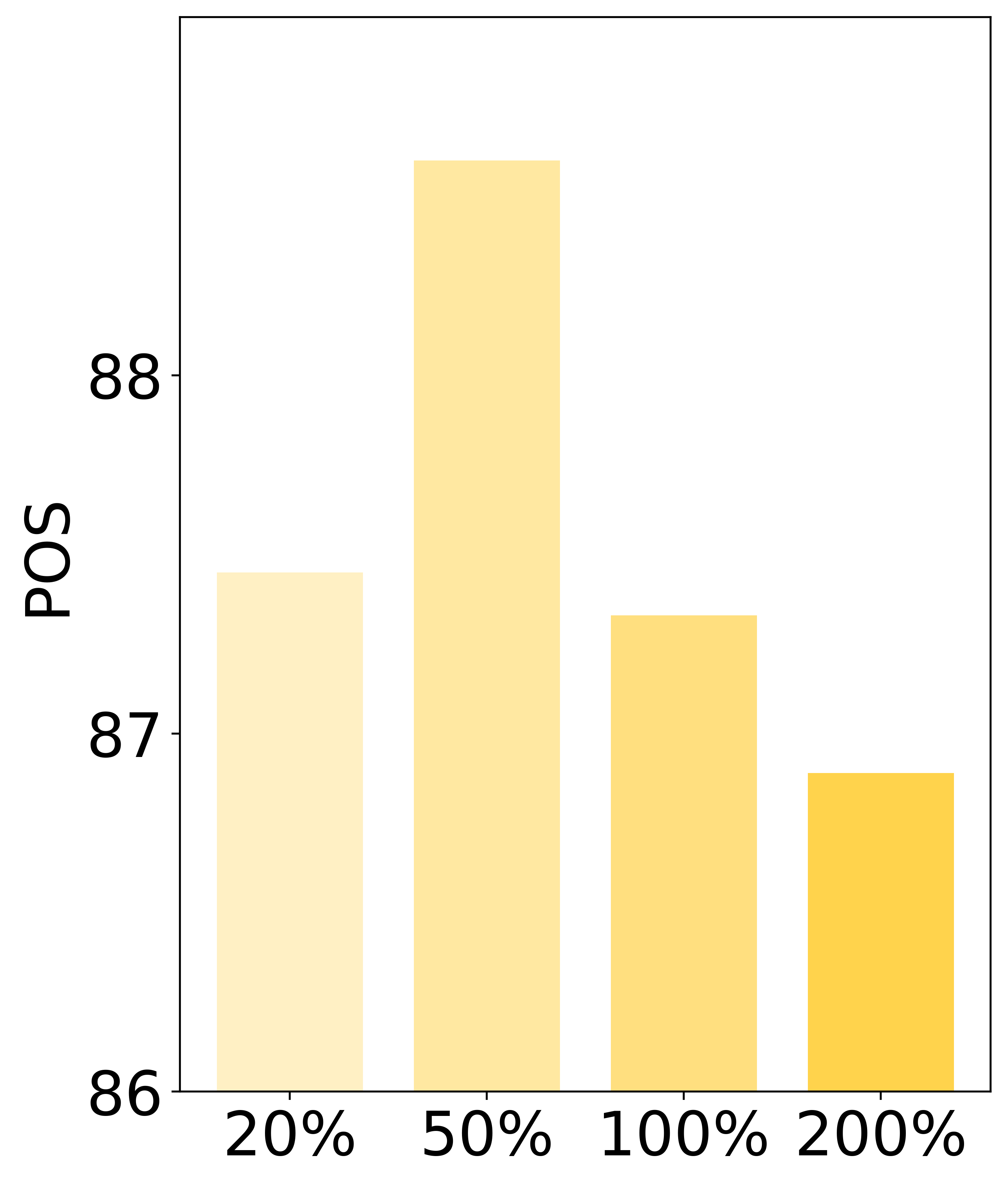}
    \end{subfigure}
    \caption{ Experimental results on the WebSRC dev set while introducing different levels of visual noise in VMD.}
    \label{fig:visual_noise}
\end{figure}

\subsubsection{Tree Structure Prediction.} 
\begin{table}[t]
    \caption{Ablation study of different input features on Tree Structure Prediction.}
    \centering
    \begin{tabular}{p{0.3cm}<{\centering}@{-}lc}
         \toprule
         \multicolumn{2}{c}{Method} & TSP Accuracy \\
        \midrule

         \multicolumn{2}{l}{$\text{WebLM}_{\text{BASE}}$} & 99.44 \\
         &  w/o closing tag & $\text{96.54}$ \\
         & w/o closing tag \& tag order & $\text{95.38}$ \\
         & w/o visual feature & $\text{90.18}$ \\
         & w/o visual feature \& closing tag   \& tag order & 72.38 \\
         \bottomrule
    \end{tabular}
    \label{tab:tsp_feature}
\end{table}
We also observe the contribution of different parts of the WebLM input to the prediction of HTML structure. By setting aside a portion of the test data, we tested the accuracy of the TSP task on this test set after training the model with different settings for 10,000 steps. "W/o Closing tag" refers to the situation where we have removed all closing tags from the HTML structure tokens. "w/o tag order" refers to the condition where, after removing these closing tags, we shuffle all structure nodes for input. These two ablation experiments are designed to observe how the model predicts the DOM Tree structure using the HTML structure input. Our results in Table \ref{tab:tsp_feature} show that different features all contribute to predicting HTML structure and complement each other. Notably, visual feature plays a crucial role in modeling HTML structure as shown in the table.

\subsection{Impact of Various Noise Levels in VMD}
Figure \ref{fig:visual_noise} shows the impact of the VMD pre-training task when different levels of noise are applied to the images. Our results show that the model performs best when the noise is either enlarging or reducing the image region by 50\%. We believe that when the noise is too small, it does not help the model to learn the robustness of visual information and alignment of visual and textual information, whereas when the noise is too large, it interferes with the model's understanding and learning of visual features.

\section{Conclusion}
In this work, we address the automated webpage understanding and information extraction by incorporating hierarchical visual information through multimodal pre-training. We primarily leverage the structured correspondence between HTML code and corresponding webpage screenshots to construct input for WebLM and perform information fusion across different modalities by devising pre-training tasks. Extensive experiments demonstrate the effectiveness of the proposed architecture, and subsequent ablation studies further highlight the importance of visual information in the process of webpage understanding. In the future, we plan to apply the WebLM to scanned/digital-born documents. By conducting automated analysis and structure construction on these documents, we aim to address the hierarchical alignment problem between image and text modalities in such document scenarios.

\section*{ACKNOWLEDGMENTS}
This work is funded by the China NSFC Projects (92370206, U23B2057, 62106142, and 62120106006) and Shanghai Municipal Science and Technology Major Project (2021SHZDZX0102).

\bibliographystyle{ACM-Reference-Format}
\balance
\bibliography{sample-base}

%









\end{document}